\documentclass[10pt,twocolumn,letterpaper]{article}

\usepackage{iccv}              

\definecolor{revBlue}{RGB}{77,139,183}

\newcommand{\datasename}{mmABC}

\definecolor{iccvblue}{rgb}{0.21,0.49,0.74}
\usepackage[pagebackref,breaklinks,colorlinks,allcolors=iccvblue]{hyperref}
\usepackage{multirow}
\usepackage{floatrow}
\usepackage{subcaption}
\newcommand{\methodname}{CMT}
\newcommand{\datasetname}{mmABC}
\newcommand*{\affaddr}[1]{#1} 
\newcommand*{\affmark}[1][*]{\textsuperscript{#1}}

\def\paperID{xxx} 
\def\confName{ICCV}
\def\confYear{2025}

\title{CMT: A \underline{C}ascade \underline{M}AR with \underline{T}opology Predictor for \\ Multimodal Conditional CAD Generation}

\author{
\bf
Jianyu Wu\affmark[1,2]~~~ 
Yizhou Wang\affmark[3]~~~ 
Xiangyu Yue\affmark[3]~~~ 
Xinzhu Ma\affmark[1]~~~ 
Jinyang Guo\affmark[4]~~~ \\
\bf
Dongzhan Zhou\affmark[1]~~~
Wanli Ouyang\affmark[1,3]~~~
Shixiang Tang\affmark[1]\textsuperscript{$\dag$}~~~ \\
\affaddr{\affmark[1]Shanghai Artificial Intelligence Laboratory~~~}
\affaddr{\affmark[2]Shanghai Jiao Tong University} \\
\affaddr{\affmark[3]The Chinese University of Hong Kong~~~}
\affaddr{\affmark[4]Beihang University~~~}\\
{\tt\small \{wujianyu,tangshixiang\}@pjlab.org.cn}
$^\dag$Corresponding author
}

\begin{document}

\twocolumn[{%
\renewcommand\twocolumn[1][]{#1}%
\maketitle
\begin{center}
    \centering
    \captionsetup{type=figure}
    \includegraphics[width=0.88\linewidth]{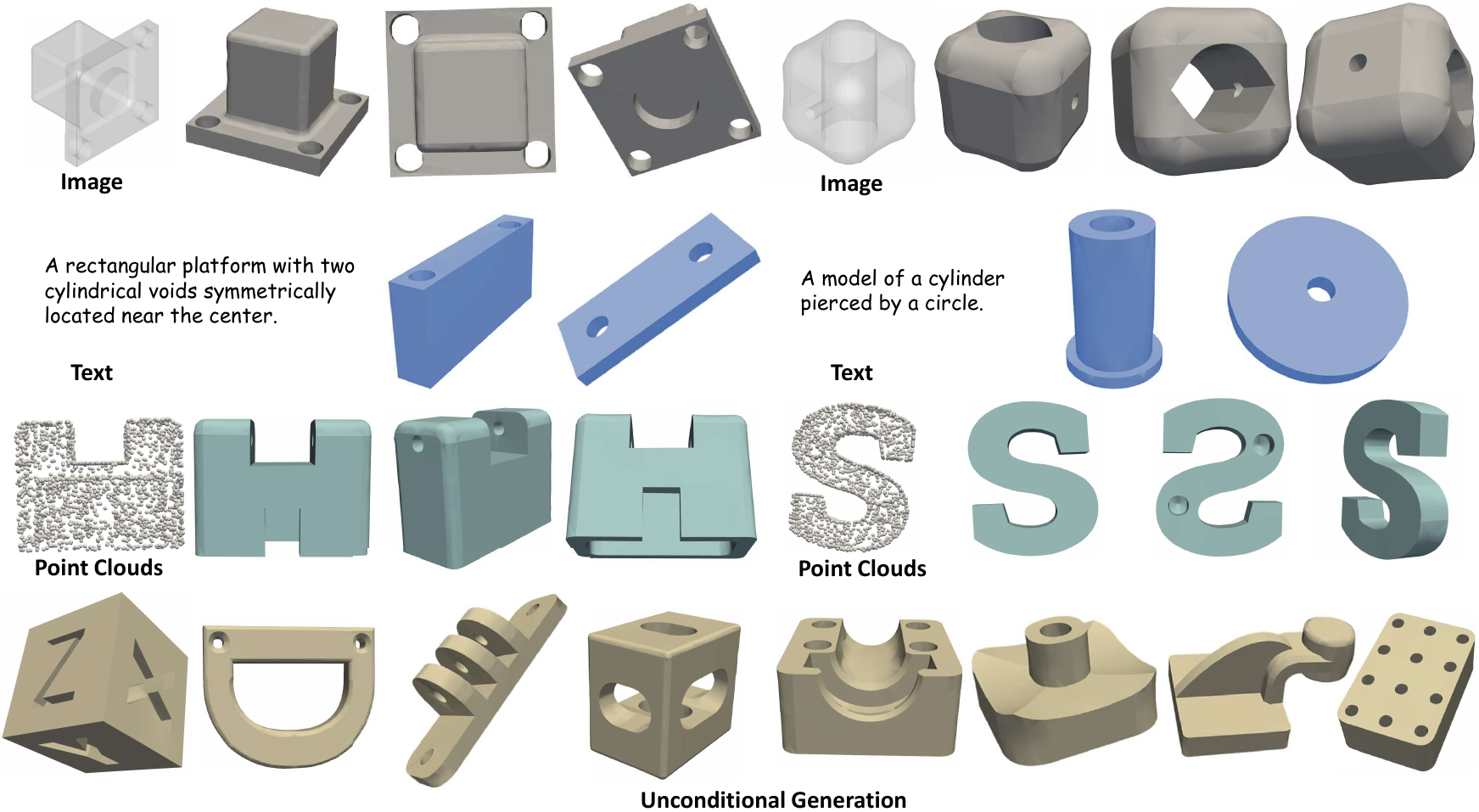}
    \captionof{figure}{\textbf{\methodname{} generates accurate and diverse CAD models} of \emph{Boundary Represention} based on user's multimodal inputs, including texts, point clouds and mulitview images, which can be seamless applied in the design and industrial manufacturing. 
    }
    \label{fig:teaser}
\end{center}%
}]

\begin{abstract}
While accurate and user-friendly Computer-Aided Design (CAD) is crucial for industrial design and manufacturing, existing methods still struggle to achieve this due to their over-simplified representations or architectures incapable of supporting multimodal design requirements. In this paper, we attempt to tackle this problem from both methods and datasets aspects. First, we propose a \underline{c}ascade \underline{M}AR~\cite{li2024autoregressive} with \underline{t}opology predictor (\textbf{\methodname{}}), the first multimodal framework for CAD generation based on Boundary Representation (B-Rep). Specifically, the cascade MAR can effectively capture the ``edge-counters-surface'' priors that are essential in B-Reps, while the topology predictor directly estimates topology in B-Reps from the compact tokens in MAR. Second, to facilitate large-scale training, we develop a large-scale multimodal CAD dataset, \textbf{\datasetname{}}, which includes over 1.3 million B-Rep models with multimodal annotations, including point clouds, text descriptions, and multi-view images. Extensive experiments show the superior of  \methodname{} in both conditional and unconditional CAD generation tasks. For example, we improve Coverage and Valid ratio by \textbf{+10.68\%} and \textbf{+10.3\%}, respectively, compared to state-of-the-art methods on ABC~\cite{koch2019abc} in unconditional generation.  \methodname{} also improves \textbf{+4.01} Chamfer on image conditioned CAD generation on \datasename{}.
\end{abstract}    
\section{Introduction}
\label{sec:intro}


Computer-Aided Design (CAD) is to use computers to aid the creation, modification, and optimization of nearly every man-made objects, playing a pivotal role in architecture and industrial design \cite{castellino2005computer,rapp2021mlcad,liu2025afbench}. Traditional CAD workflows before the era of deep learning often involve multiple processes, including 2D sketches (\emph{e.g.}, circles, lines, splines), 3D operations (\emph{e.g.}, extrusion, loft, fillet) of these 2D elements, with the resulting CAD model ultimately being converted into a B-Rep~\cite{lambourne2021brepnet} format, making the whole design process time-consuming and costly. Experts are required to repeat the process to design tailored CAD models according to the multimodal users requirements, either by textual descriptions or visual clues, \emph{e.g.}, images and point clouds. Therefore, an ideal CAD generation tool should
not only generate topologically and geometrically accurate models but also support multimodal conditioning while remaining user-friendly, even for non-experts.

With the advancements of deep learning~\cite{he2016deep,ren2016faster,su2020adapting,li2024autoregressive}, recent methods attempt to leverage large generative networks to empower fully autonomous CAD generation~\cite{xu2022skexgen,xu2023hierarchical,alam2024gencad}. These deep-learning-based tools can be broadly categorized into \emph{MLLM-based methods} \cite{wu2021deepcad,khan2024cad,jayaramansolidgen,xu2024cad} and \emph{diffusion-based methods} \cite{xu2024brepgen}, but they both fail to simultaneously process multimodal inputs and generate accurate models. \emph{MLLM-based methods} model CAD as a sequence of discrete tokens, \emph{i.e.}, operating commands or predefined geometry templates, and then generate them token-by-token. Combined with multimodal data, \emph{MLLM-based methods} can generate CAD models by conditioning on texts \cite{khan2024text2cad,badagabettu2024query2cad}, images \cite{chen2024img2cad,alam2024gencad}, point clouds \cite{uy2022point2cyl,dupont2024transcad}, \emph{etc}. However, such approaches are limited in accurately generating complex CAD models, \emph{e.g.}, chamfers, fillets, and freeform surfaces, due to limited representation capability of discrete tokens. Specifically, command sequential tokens only include sketch and extrusion operation, which fails to model shapes that require other commands, and shape template tokens restrict design space of CAD models, which fails to generate complex CAD models out of templates. \emph{Diffusion-based} methods can generate B-Rep models in a continuous fashion~\cite{ho2020denoising}, yet struggle to handle generation tasks with multimodal conditions, which is not user-friendly to CAD engineers. To achieve both generation quality with availability to users's multimodal instructions, we aim to develop a novel method to generate CAD models in continuous space while ensure both precise topology and strong compatibility with multimodal conditioning.

Masked autoregressive network (MAR), recently proposed by Kaiming He~\cite{li2024autoregressive}, 
unifies autoregressive token interdependence modeling with diffusion-based per-token distribution learning, opening new frontiers in image generation.
While this architecture offers opportunities of accurate and multimodal CAD generation by autoregressively decoding tokens to continuous values, 
the vanilla MAR lacks inherent mechanisms to capture the essential topological relationships (\emph{e.g.}, vertices-edge-surface hierarchies) in B-Reps.
In this paper, we devote substantial efforts to alleviating this challenge by two innovative designs, \emph{i.e., the cascade autoregressive networks} and \emph{the topology predictor module.} The cascade autoregressive diffusion network designs an edge MAR followed by a surface MAR which leverages edge MAR outputs as its conditions, following the topology priors of B-Reps, \emph{i.e.}, edges contour a surface. The topology predictor module can directly predict the topology from the compact latent codes, which is over 4200 times faster than the post-processing algorithm in~\cite{liu2024point2cad}\footnote{The time for topology prediction of 256 CAD models by post-processing in Point2CAD~\cite{liu2024point2cad} (CVPR 2024) is 161.15s while our predictor costs 0.038s.}. Integrating such designs with a unified multimodal condition encoder, we can generate CAD conditioned on multimodal inputs by appending multimodal tokes as start of sequential tokens in autoregressive networks, and preserve the accuracy by diffusively decoding tokens into continuous point, edge and surface representations.

Accordingly, we introduce \methodname{}, a \textbf{C}ascade \textbf{M}AR with \textbf{T}opology predictor for multimodal conditional CAD generation, which is the first multimodal CAD generation framework that can generate complex geometry and topology based on B-Rep.
\methodname{} consists of a unified multimodal condition encoder, a cascade autogressive generation network for generating the edges and surfaces, respectively, and a topology prediction module~\cite{vaswani2017attention} for predicting the topological relationships between the generated edges and surfaces. Given multimodal conditions, the encoder extract their embeddings, which will be fed into the cascade autoregressive network to generate latent tokens
in a ``edges-then-surfaces'' manner.
These latent tokens are afterwards decoded by a denoising network~\cite{ho2020denoising} to accurate representation of edges and surfaces. By supervising the topology of the generated edges and surfaces, our proposed \methodname{} can generate accurate and complex CAD models with multimodal conditions.

To fully unleash the potential of our \methodname{}, we further standardize and annotate the existing CAD dataset \cite{koch2019abc}, resulting in the largest datasets \emph{\textbf{\datasename{}}} containing over 1.3 million CAD models with multimodal annotations, \emph{i.e.}, multi-view images, textual descriptions and point clouds. By training on this dataset, our \methodname{} showcases strong unconditional CAD generation capabilities and conditional CAD generation capabilities conditioned on multimodal cues. Specifically, in unconditional generation tasks, our Coverage increased by \textbf{+10.68\%}, and the Valid ratio improved by \textbf{+10.3\%} compared to the previous state-of-the-art~\cite{xu2024brepgen} on ABC~\cite{koch2019abc}. In conditional generation tasks, \methodname{} also achieves remarkable and consistent improvement, \emph{e.g.}, \textbf{+0.13} Chamfer on point cloud conditioned CAD generation, \textbf{+4.01} Chamfer on image conditioned CAD generation on our proposed \datasename{}.

To conclude, our contributions are three-fold: (1) We introduce \methodname{}, the first multimodal CAD generation framework based on B-Rep representation, enabling both unconditional generation and condition generation from point cloud, multi-view images, and text descriptions; (2) We propose the cascade autoregressive network and topology predictor module to tackle the most important challenge of B-Rep generation, \emph{i.e.}, topology of B-Reps. (3) We further create \datasename{}, the largest multimodal dataset for CAD generation, which consists of over 1.3 million B-Rep models paired with point clouds, text descriptions, and multi-view images annotations. 



\section{Related Work}
\label{sec:relate}

\subsection{CAD Generation}

\noindent\textbf{MLLM-based methods}. Inspired by the success of MLLM, some methods attempt to generate CAD using MLLM architecture, viewing CAD as an iterative command sequence for autoregressive generation \cite{wu2021deepcad,xu2023hierarchical,khan2024cad,jayaramansolidgen,xu2024cad}. For instance, DeepCAD~\cite{wu2021deepcad} employs a transformer as a decoder to sequentially decode lexicalized commands, which are then converted into editable CAD files through geometric kernels post-processing. CAD-MLLM~\cite{xu2024cad} is built upon a pretrained multimodal large language model \cite{zhu2023minigpt,wu2024next,liu2023visual,li2023blip}, finetuned to enable the generation of command sequences. SolidGen~\cite{jayaramansolidgen} uses predefined templates and discretization \cite{van2017neural,razavi2019generating,guo2024compressing} to achieve autoregressive generation. While MLLM-based methods integrate easily with existing frameworks, current sequential CAD generative networks are still constrained to generating sketches composed of basic primitives such as lines, arcs, and circles, and are limited to the extrude modeling operations.

\noindent\textbf{Diffusion-based methods}. Other methods choose to use diffusion for denoising a continuous CAD representation~\cite{alam2024gencad,xu2024brepgen}. For example, BrepGen~\cite{xu2024brepgen} completes B-Reps to the same length through node duplication, and successively generates all the surfaces, edges and vertices at one time through diffusion~\cite{ho2020denoising}. Although these methods can generate plausible B-Reps, diffusion-based methods struggle  to accurately generate the corresponding model according to the user's multimodal input. Therefore, we propose to use user-friendly cascade autoregressive generation network to gradually generate B-Rep while keep the accurate and continuous characteristics of B-Rep.

\subsection{CAD Datasets}
Various CAD datasets have been proposed to promote the development of CAD generation \cite{koch2019abc,wu2021deepcad,dupont2022cadops,zhou2023cadparser,willis2021fusion,seff2020sketchgraphs,li2022free2cad,yuan2024openecad,chen2024img2cad,you2024img2cad,xu2024cad}. However, the existing datasets are either on a small scale, or lacking multimodal data pairs, or based on command sequence that is shortage of diversity and accuracy. As shown in \cref{tab:datasets}, the largest CAD dataset, ABC~\cite{koch2019abc}, while contains 1 million CAD models, there is a lack of multimodal inputs corresponding to each CAD model. Although datasets such as Free2CAD~\cite{li2022free2cad} and Text2CAD~\cite{khan2024text2cad} include modal input, the dataset size remain relatively small and the data modality is single. The recently proposed command sequence-based dataset Omni-CAD~\cite{xu2024cad}, annotates multimodal data for the ABC dataset. However, it is based on command sequence, resulting in lacks in diversity and accuracy. In order to address the absence of existing datasets and to promote the development of future CAD research, especially B-Rep generation, we construct \datasetname{}, a large-scale multimodal dataset based on B-Rep. To the best of our knowledge, our \datasetname{} dataset is the largest multimodal CAD dataset to date, containing over 1.3 million CAD models and their corresponding multimodal inputs.




\begin{table}[t]
\centering
\resizebox{0.99\linewidth}{!}{
\begin{tabular}{ccccccc}
\toprule
Dataset     &  Rep. &  Dataset Size  & Image & Point & Text & FM \\ 
\midrule
ABC~\cite{koch2019abc}  &	B-Rep	& \dag1,000,000 &      &      &      &   $\checkmark$  \\
CC3D-Ops~\cite{dupont2022cadops}  & B-Rep & $\sim$37,000 &      &      &      &   $\checkmark$ \\
DeepCAD~\cite{wu2021deepcad}  & CS & 179,133 &      &      &      &    \\
Free2CAD~\cite{li2022free2cad}	& CS	& $\sim$210,000	&   $\checkmark$   &      &      &     \\
Text2CAD~\cite{{khan2024text2cad}}   & CS & $\sim$158,000 &      &      &   $\checkmark$   &   \\
Omni-CAD~\cite{xu2024cad} & CS & 453,220 &   $\checkmark$   &   $\checkmark$   &  $\checkmark$    &     \\
\textbf{mmABC (Ours)} &	B-Rep	& \textbf{1,353,251} &  $\checkmark$    &   $\checkmark$   &  $\checkmark$    &   $\checkmark$  \\
\bottomrule
\end{tabular}
}
\caption{Comparison of previous datasets and our \datasetname{}. B-Rep and CS represent boundary representation and command sequence, respectively. Rep: Representation. FM: Freeform modeling denotes supporting arbitrary surfaces and edges. \dag: Removing similar models and multi-body models in ABC, which are not beneficial for training, leads to about 0.6 million CAD models. } 
\label{tab:datasets}
\end{table}

\section{mmABC: a Multimodal B-Rep Dataset}
\label{sec:dataset}

To enable accurately generate B-Rep models based on given multimodal instructions, a large-scale multimodal dataset is required. However, as shown in \cref{tab:datasets}, the existing datasets are either small in size \cite{dupont2022cadops,wu2021deepcad,li2022free2cad,khan2024text2cad,xu2024cad}, or lack multimodal inputs \cite{koch2019abc,dupont2022cadops,wu2021deepcad}, or are based on command sequence that do not support freeform modeling \cite{wu2021deepcad,li2022free2cad,khan2024text2cad,xu2024cad}. To tackle the problem, we introduce \datasetname{}, the largest CAD dataset to date, including over 1.3 million B-Rep models and their multimodal descriptions, \emph{i.e.}, multi-view images, point clouds, and text descriptions, as depicted in Fig.~\ref{fig:teaser}. 


\begin{figure*}[t]
  \centering
   \includegraphics[width=0.95\linewidth]{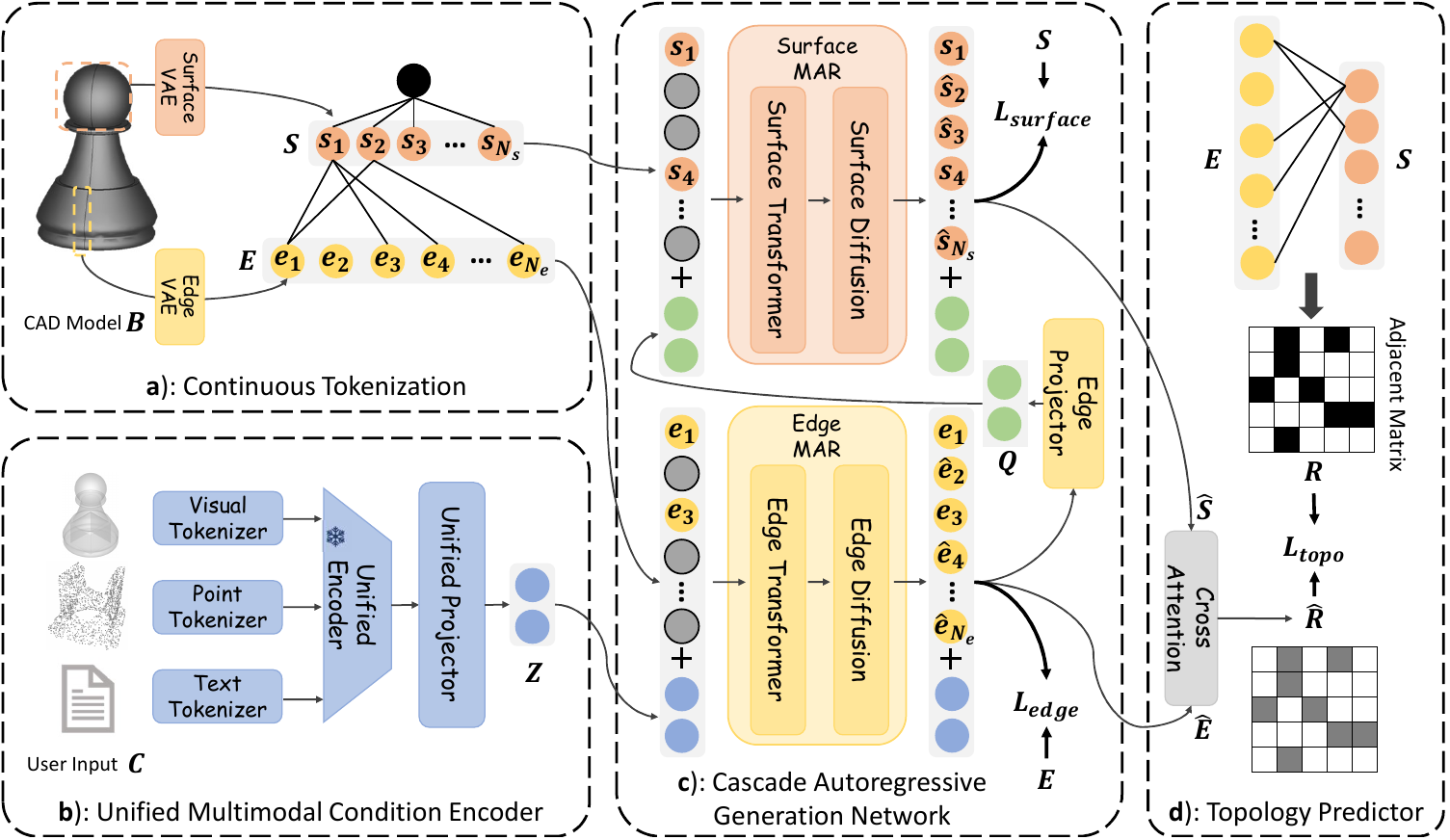}
   \caption{\textbf{The pipeline of \methodname{}}. \methodname{} consists of continuous tokenization of B-Rep, a unified multimodal condition encoder, a cascade autoregressive generation network, and a topology predictor. Given a CAD model $B$ and its corresponding multimodal inputs $C$, (a) the continuous tokenization process in \methodname{} converts the B-Rep into continuous surface tokens $S$ and edge tokens $E$ using Surface VAE~\cite{kingma2013auto} and Edge VAE respectively. (b) the users' input $C$ is encoded into a fixed-length condition embedding $Z$ in the unified multimodal condition encoder. (c) The edge sequence is randomly masked, and the edge MAR is trained to reconstruct it from learnable masked tokens $M_{e}$ based on visible edge tokens $E_v$ and the condition embedding $Z$. The generated varying‑length edge sequence, along with the condition embedding, is further encoded into a fixed-length edge embedding $Q$, which are concatenated with the visible surface tokens $S_v$ and masked tokens $M_{s}$ to reconstruct masked surface tokens. (d) a cross-attention layer~\cite{vaswani2017attention} is utilized to predict the adjacent matrix $\hat{R}$ between the generated edge tokens $\hat{E}$ and the generated surface tokens ${\hat{E}}$. The overall loss comprises the topology error $L_{topo}$ from the adjacency matrix prediction, as well as the token generation loss of edges $L_{edge}$ and surfaces $L_{surf}$.}
   \label{fig:overview}
\end{figure*}

\noindent\textbf{Dataset construction.} mmABC is constructed upon the widely-used ABC dataset~\cite{koch2019abc}. Nevertheless, directly annotating ABC with multimodal descriptions is not enough, because ABC contains many identical and multi-body CAD models, which will potentially hamper the network training. To address these problems, we decompose complex multi-body models into multiple basic single models to augment the dataset. To identify and remove similar models, we filtere data with same hash value of their 6-bit quantized point coordinates sampled from surfaces of CAD models, obtaining the constructed mmABC dataset with 1,353,251 high-quality CAD models. 

\noindent\textbf{Annotation generation.} (a) Multi-view images: We use OpenCASCADE~\cite{banovic2018algorithmic} to render 14 fixed-angle images for each B-Rep model. (b) Point clouds: We take points sampling on each CAD surface with its corresponding normal vectors as point clouds descriptions. (c) Text: To generate a textual description of the model, we utilize rendered multi-view images to generate the caption of the model with Vision-Language Models (VLMs). Specifically, we employ the open sourced InternVL2-40B~\cite{chen2024internvl} and randomly input two or more views into it, with randomly selected prompts from the pre-designed prompt templates, so as to generate diverse and high quality text descriptions. More examples and details are shown in the supplementary materials. 
\section{\methodname{}}
\label{sec:method}

In order to generate complex CAD models from multimodal inputs, we design a cascade autoregressive generation network, tailored to B-Rep topology, to progressively generate B-Rep in an ``edge-then-surface" manner. To empower vanilla MAR to capture the topology of B-Reps, our innovations lie in the ``edge-to-surface'' cascade architecture and the topology predictor.
Our pipeline is described below:

\textit{Step 1}: Given a B-Rep model $B$, we convert it into a continuous edge sequence $E$ and a continuous surface sequence $S$ to facilitate the autoregressive generation while enabling accurate CAD modeling (\cref{sec:tokenization}) . 

\textit{Step 2}: The multimodal user inputs $C$ are encoded by the \emph{Unified Multimodal Condition Encoder} $\mathcal{F}$ into fixed-length condition embedding $Z$ (\cref{sec:encoder}). 

\textit{Step 3}: Based on condition embedding $Z$, 
\emph{Cascade Autoregressive Network} generate edges and surfaces in an ``edge-to-surface'' manner (\cref{sec:backbone}). Specifically, the edge transformer $\mathcal{G}_e$ generates edge sequences $E$ from input $Z$ and the surface transformer $\mathcal{G}_s$ generates a surface sequence $S$ based on $Z$ and generated edge sequences $E$.

\textit{Step 4}: Predict the topology relationship $R$ using the generated edge sequence $E$ and surface sequence $S$ with \emph{Topology Predictor} $\mathcal{H}$ to associate each surface with its corresponding surrounding edges, ultimately yielding a complete B-Rep of the model (\cref{sec:predictor}).

\subsection{Continuous Tokenization}
\label{sec:tokenization}



To avoid the loss of precision and diversity caused by discrete representations and facilitate autoregressive training, we use continuous tokenization to convert the primitives (surfaces, edges and vertices) of B-Rep into the respective sequence of continuous-value tokens, as shown in Fig.~\ref{fig:overview} (a). 


\noindent\textbf{Surface token.} Given a surface, we first sample points uniformly from the normalized surface as its geometric information and use a Surface VAE~\cite{kingma2013auto} to encode it into the hidden codes. Furthermore, motivated by BrepGen~\cite{xu2024brepgen}, we utilize the bounding box enclosing each surface to represent its topology characteristics. The bounding box coordinates and hidden features are then concatenated as a surface token $s$ for the given surface, which effectively combines both geometric and topological information. 

\noindent\textbf{Edge token.}
We combine the information of the edge and its two adjacent vertices (start point and end point) together to the edge token $e$. Concretely, given an edge, we first extract its hidden features with an Edge VAE~\cite{kingma2013auto}, obtain the coordinates of its bounding box and adjacent vertices. Then we concatenate these elements to form the edge token $e$. Note that the vertices information is integrated in the edge tokens, there is no need to generate vertice tokens.

\noindent\textbf{Token ordering.}
After tokenization, we order surface tokens and edge tokens respectively to enable autoregressive training. Following common practice in mesh generation~\cite{siddiqui2024meshgpt,chen2024meshxl}, we sort tokens in ascending order based on the 3D coordinate value of $x_{1}, y_{1}, z_{1}, x_{2}, y_{2}, z_{2}$ within the bounding boxes. This process yields an ordered surface sequence $S=\{s_1, s_2,...,s_{N_s}\}$ and edge sequence $E=\{e_1, e_2,...,e_{N_e}\}$, where $N_s$ and $N_e$ denote the number of surfaces and edges in the B-Rep model $B$, respectively.

\subsection{Unified Multimodal Condition Encoder}
\label{sec:encoder}

To prevent redundant training of modality-specific condition encoders,
we design a Unified Multimodal Condition Encoder $\mathcal{F}$ to encode them into fixed-length condition embedding, \emph{i.e.}, \(Z = \mathcal{F}(C)\), where $C$ are multimodal user inputs and $Z$ denotes condition embedding.
Inspired by OneLLM~\cite{han2024onellm}, our Unified Multimodal Condition Encoder consists of modality-specific tokenizers, a unified encoder and a unified projector as depicted in \cref{fig:overview} (b). Specifically, given mulitmodal conditions $C$, we first use modality-specific tokenizers to transform them into token sequences, \emph{i.e.}, CLIP tokenizers~\cite{radford2021learning} for image and text, 3D convolution layer for point clouds, followed by a frozen CLIP-ViT~\cite{radford2021learning} encoder as the unified encoder to extract the condition features from the tokens. Afterwards, we leverage a learnable projector to extract condition embedding $Z$. 

\subsection{Cascade Autoregressive Network}
\label{sec:backbone}
While MLLM-based and diffusion-based methods can not simultaneously accept multimodal conditions and generate complex models, the newly proposed MAR~\cite{li2024autoregressive} offers the opportunity to take advantages of both frameworks. To modify the vanilla MAR suitable for CAD generation, we design a Cascade Autoregressive Generation Network (CAN) to embed the topology prior of B-Reps, \emph{i.e.}, edges contour surfaces. Specifically, 
the Cascade Autoregressive Network consists of two MARs~\cite{li2024autoregressive}, \emph{i.e.}, Surface MAR and Edge MAR. Each MAR contains a transformer~\cite{vaswani2017attention} $\mathcal{G}$ and a diffusion~\cite{ho2020denoising} MLP network $\mathcal{D}$. In practice, we divide the B-Rep generation process into two sequential steps: edge token generation followed by surface token generation.

\noindent\textbf{Edge token generation.} Inspired by MAR~\cite{li2024autoregressive}, given the sequence of edge tokens $E$, we mask several tokens and replace them with learnable masked tokens $M_{e}$ using a randomly sampled mask $m$.
To reconstruct masked edge tokens $E_{m}$, we feed the learnable masked tokens $M_{e}$, condition embedding $Z$ and the remaining visible edge tokens $E_{v}$ into the edge transformer $\mathcal{G}_e$ to obtain the features of masked edge tokens $c_e$ by $c_e=\mathcal{G}_e(Z,E_v,M_e)$. The features then serve as the condition of edge diffusion network $\mathcal{D}_e$ to denoise edge tokens $\hat{E}_{m}$ \emph{i.e.}, $\hat{E}_{m} = \mathcal{D}_e(c_e)$.

\noindent\textbf{Surface token generation:}
In surface token generation, \methodname{} leverages both conditional embedding $Z$ and generated edge tokens $E$ to generate surface tokens $S$.
Since different models have varying numbers of edges, we introduce an edge projector composed of self-attention layers and learnable edge tokens to extract fixed-length edge condition embedding $Q$. 
Similar to edge token generation, the surface sequence $S$ is randomly masked and replaced by learnable mask tokens $M_s$. The edge condition embedding $Q$, visible surface tokens $S_v$ and mask tokens $M_s$ are fed into the surface transformer $\mathcal{G}_s$ to produce the surface features $c_s$ through $c_s=\mathcal{G}_s(Q,S_v,M_s)$. The masked surface tokens $\hat{S}_{m}$ can be afterwards denoised by surface features $c_s$ and surface diffusion network $\mathcal{D}_s$, \emph{i.e.}, $\hat{S}_{m} = \mathcal{D}_s(c_s)$.

\subsection{Topology Predictor}
\label{sec:predictor}

In order to reconstruct the final B-Rep, we aim to obtain the topology relationship $R\in \mathbb{R}^{N_e\times N_s}$ between the generated edge tokens $\hat{E}$ and surface tokens $\hat{S}$ through Topology Predictor. Thanks to the precise generation of edge tokens and surface tokens, we directly predict topology relationship $\hat{R}$ between edges and surfaces with a simple cross-attention layer~\cite{vaswani2017attention} $\mathcal{G}_t$ \emph{i.e.}, $A = \mathcal{G}_t(\hat{E}, \hat{S})$. A threshold $\tau$ is set to determine whether a relationship exists. When the score in the generated matrix $A \in \mathbb{R}^{N_e\times N_s}$ exceeds the threshold $\tau=0.5$, it is considered that there is an adjacency relationship between the edge and the surface.

\subsection{Objective Function}
\label{sec:objective}
The objective function includes two parts: token generation and topology prediction. Similar to MAR, we generate the conditions $c_e,c_s$ of the subsequent diffusion~\cite{ho2020denoising} in an autoregressive way, then the objective function of token generation becomes a constraint on the error of the noise predicted by diffusion $D_e,D_s$:
\begin{equation} 
    L_{edge}(e,c_e) = \mathbb{E}_{t,\, \epsilon} \Bigl[ \bigl\|e - D_e(e_t,|t, c_e) \bigr\|^2 \Bigr], e\in E_m,
\end{equation}
\begin{equation} 
    L_{surf}(s,c_s) = \mathbb{E}_{t,\, \epsilon} \Bigl[ \bigl\|s - D_s(s_t,|t, c_s) \bigr\|^2 \Bigr], s\in S_m,
\end{equation}
where $e_t\!=\!\sqrt{\bar{\alpha}_t}\,e + \sqrt{1-\bar{\alpha}_t}\,\epsilon$ and $s_t=\sqrt{\bar{\alpha}_t}\,s + \sqrt{1-\bar{\alpha}_t}\,\epsilon$, $\epsilon$ is sampled from $\mathcal{N}(0, \mathbf{I})$, $\alpha_t$ defines a noise schedule.

The objective function of topology prediction constrains the error between the predicted matrix $A$ and the target adjacent matrix $R$:
\begin{equation} 
    L_{topo}(A,R) = \frac{1}{N_e\times N_s}\sum_{i=1}^{N_e}\sum_{j=1}^{N_s} \left( A_{ij} - R_{ij} \right)^2. 
\end{equation}
Therefore, the total objective function of \methodname{} is:
\begin{equation}
    L = L_{edge} + L_{surf} + L_{topo}.
\end{equation}





\section{Experiment}
\label{sec:experiment}

\begin{figure*}[t]
  \centering
   \includegraphics[width=0.99\linewidth]{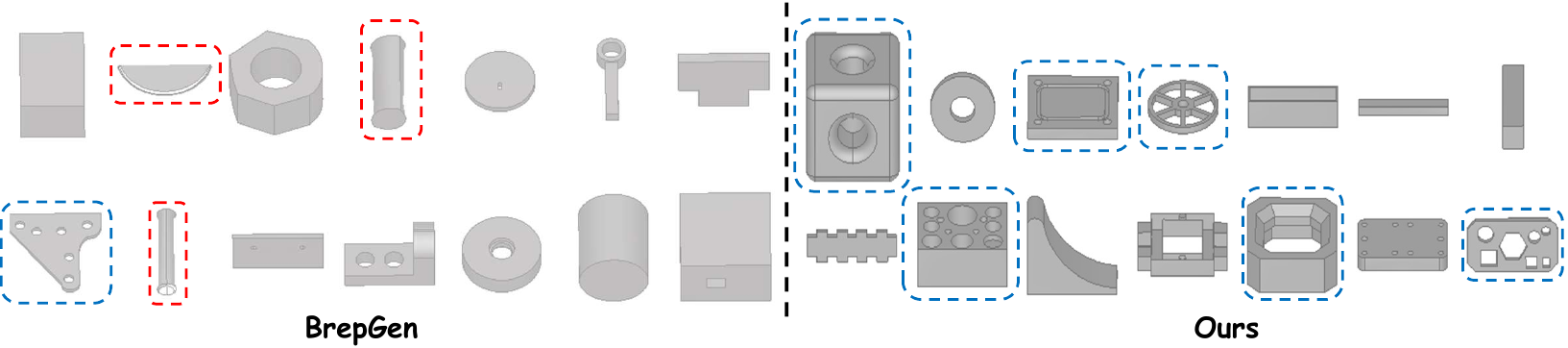}
\vspace{-0.5em}
   \caption{Unconditional generation results on ABC~\cite{koch2019abc} dataset by BrepGen~\cite{xu2024brepgen} and our \methodname{}. The blue rectangles represent complex models (with more than 20 surfaces after splitting), and the red rectangles represent broken models. Our method generates more complex CAD models with fewer broken cases.}
   \label{fig:uncond_fig}
\end{figure*}

\subsection{Experimental Setup}
\label{sec:setup}
\noindent\textbf{Datasets}. 
To demonstrate the effectiveness of our \methodname{}, we train it on ABC~\cite{koch2019abc}, DeepCAD~\cite{wu2021deepcad} for unconditional CAD generation tasks and on our proposed \datasetname{} for conditional CAD generation tasks. Following the dataset division strategy of ABC, we split \datasetname{} into training set, val set and test set with an 8:1:1 ratio\footnote{Our test set contains all the test CAD models in ABC. For extra CAD models generated by decoupling process that do not appeal, we randomly sample 10\% 
and 10\% to put into the val set and test set, respectively.}. 
 
\noindent\textbf{Implementation Details}.
We use pretrained CLIP-ViT-Large~\cite{radford2021learning} as the frozen unified encoder. Following MAR~\cite{li2024autoregressive}, the diffusion process~\cite{ho2020denoising} adopts a cosine schedule. We predefine maximum sequence lengths of 64 and 32 for edge and surface in the DeepCAD~\cite{wu2021deepcad} dataset, and 128 and 64 for edge and surface in the ABC~\cite{koch2019abc} and \datasetname{} dataset respectively, covering over 95\% of the data in each dataset. The sizes of condition embedding and edge condition embedding are set to $\mathbb{R}^{18\times 768}$. For unconditional CAD generation, the training process comprises 2100 epochs. For conditional CAD generation, another 1000 epochs are required. All trainings are done on 8 A100 GPUs. During inference, we set the number of sampling steps to the same length as the sequence by default, that is 64/32 on DeepCAD and 128/64 on ABC and \datasetname{}, generating one token at a time. Please refer to supplementary for more details.

\noindent\textbf{Metrics}. We follow the metrics in \cite{xu2024brepgen} and \cite{xu2024cad} for Unconditional CAD Generation and Conditional CAD Generation tasks, respectively. The results in the table are multiplied by 100. Please refer to supplementary for more details. 



\subsection{Unconditional CAD Generation}
\label{sec:unconditional}

Unconditional CAD generation aims to evaluate the modeling performance of methods for prior distribution of CAD models. Following the metrics in BrepGen~\cite{xu2024brepgen}, we compare our \methodname{} with state-of-the-art methods on DeepCAD~\cite{wu2021deepcad} and ABC~\cite{koch2019abc}. 

The quantitative results are summarized in \cref{tab:uncond}. First, compared with BrepGen~\cite{xu2024brepgen},
\methodname{} showcases consistent and remarkable improvements on COV, MMD and JSD, \emph{e.g.}, \textbf{+4.45\%} and \textbf{+10.68\%} Coverage on DeepCAD and ABC, indicating that our method has superior generation quality and a closer alignment with the ground-truth distribution. Second, despite high-quality generation performance, \methodname{} is capable of producing novel and unique models, as demonstrated by the high scores in the Novel and Unique.
Third, after 4 bit quantization, the Valid ratio of \methodname{} is also better 
compared with previous methods, \emph{i.e.}, \textbf{+7.2\%} on DeepCAD and \textbf{+10.3\%} on ABC than BrepGen. The improvement on Valid illustrates that instances of broken topology occur far less frequently in the models generated by our method, demonstrating the effectiveness of cascade masked autoregressive generation. 

\cref{fig:uncond_fig}
illustrates qualitative results for unconditional generation on ABC. We see that \methodname{} generate desirable B-Reps with complex surfaces and edges (models rounded by blue rectangels).
We also observe that compared with BrepGen, the models generated by \methodname{} have more diverse topologies and fewer error cases (such as unbounded open regions or self-intersecting edges).

\begin{table}[t]
  \centering
    \setlength\tabcolsep{3.5 pt}
  \small
  \resizebox{0.99\linewidth}{!}{
    \begin{tabular}{llccc|ccc}
    \toprule
    Dataset &{Method} & COV$\uparrow$  & MMD$\downarrow$   & JSD$\downarrow$   & Novel$\uparrow$ & Unique$\uparrow$ & Valid$\uparrow$ \\
    \midrule 
    \multirow{4}[0]{*}{DeepCAD~\cite{wu2021deepcad}}  & DeepCAD~\cite{wu2021deepcad} & 65.46 & 1.29  & 1.67  & 87.4  & 89.3  & 46.1 \\ 
          & SolidGen~\cite{jayaramansolidgen} & 71.03 & 1.08  & 1.31  & 99.1  & 96.2  & 60.3 \\
          & BrepGen~\cite{xu2024brepgen} & 71.26 & 1.04  & \textbf{0.09}  & 99.8  & 99.7  & 62.9 \\
          & \methodname{} & \textbf{75.71} & \textbf{0.92}  & 1.02  & \textbf{99.9}  & \textbf{99.8}  & \textbf{70.1} \\\midrule 
    \multirow{2}[0]{*}{ABC~\cite{koch2019abc}}  & BrepGen~\cite{xu2024brepgen} & 57.92 & 1.35  & 3.69  & 99.7  & 99.4  & 48.2 \\
          & \methodname{} & \textbf{68.60}  & \textbf{1.35}  & \textbf{2.79}  & \textbf{99.7}  & \textbf{99.7}  & \textbf{58.5} \\
          \bottomrule
    \end{tabular}
    }
    \caption{Quantitative evaluations of DeepCAD and ABC unconditional generation with Coverage (COV) percentage, Minimum Matching Distance (MMD), Jensen-Shannon Divergence (JSD), and the ratios of Unique, Novel, and Valid models. Arrow ($\uparrow$ and $\downarrow$) indicate whether higher or lower values are better. Our method, \textbf{CMT}, consistently outperforms the baselines across most metrics.}
  \label{tab:uncond}
\end{table}

\begin{figure*}[t]
    \begin{floatrow}
        \ffigbox[\FBwidth]{
         \includegraphics[width=0.99\linewidth]{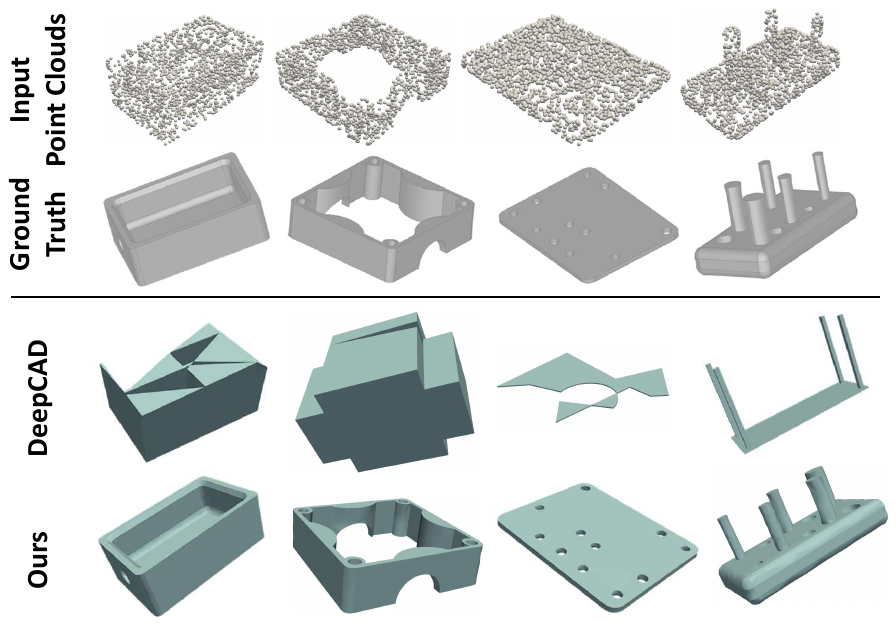}
        }{\caption{Comparisons on point cloud conditional generation.}\label{fig:pcd_cond_fig}}
        \ffigbox[\FBwidth]{
        \includegraphics[width=0.99\linewidth]{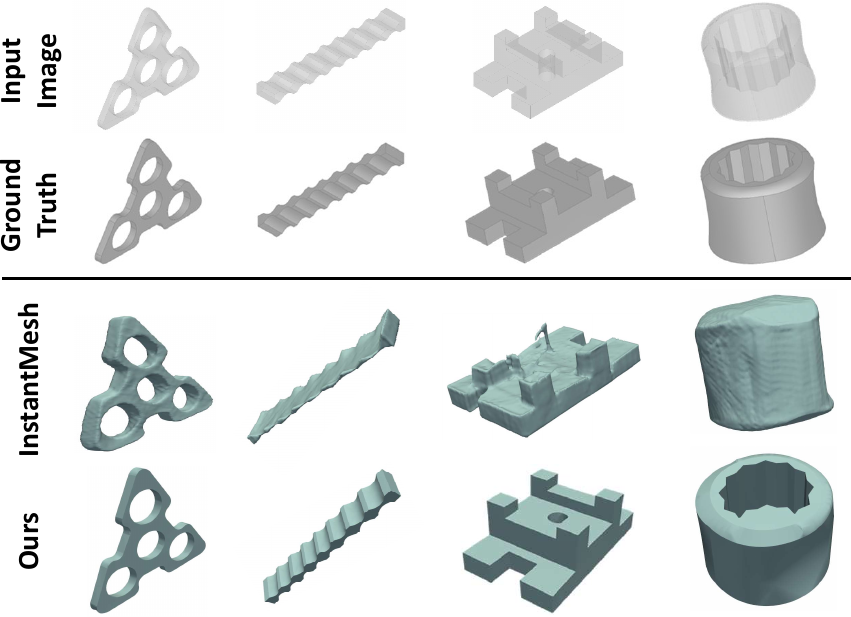}
        }{\caption{Comparisons on image-conditional generation task.}\label{fig:img_cond_fig}}
    \end{floatrow}
    \vspace{-0.5em}
\end{figure*}





\subsection{Conditional CAD Generation}
\label{sec:conditional}

Conditional CAD generation task aims to evaluate the capability of generating desirable CAD models from given multimodal instructions. These tasks are categorized according to the input modality, \emph{e.g.}, point cloud, image and text. Following the reconstruction metrics adopted in existing work \cite{liu2024point2cad,liu2024split}, we conduct experiments on point clouds, images and text, and compare \methodname{} with both reconstruction methods and generation methods.


\begin{table}
\centering \small
\setlength\tabcolsep{3.5 pt}
\small
\resizebox{0.99\linewidth}{!}{
\begin{tabular}{lccc}
\toprule
Method  & Chamfer $\downarrow$       & F-score $\uparrow$ & Normal C $\uparrow$ \\
\midrule
\multicolumn{4}{l}{\textit{Reconstruction methods}}       \\
\midrule
ComplexGen~\cite{guo2022complexgen} & 2.04          & 94.81          & 71.01     \\
Point2CAD~\cite{liu2024point2cad}    & 1.55          & 95.83          & 77.24     \\

NVDNet~\cite{liu2024split}    & 0.77          & 98.17          & 94.36     \\
\midrule
\multicolumn{3}{l}{\textit{Generation methods}}       \\
\midrule
DeepCAD~\cite{wu2021deepcad}    & 5.11          & 83.56          & 69.58     \\
\methodname{}       & \textbf{0.64}         &  \textbf{99.07}          & \textbf{95.48}     \\
\bottomrule
\end{tabular}
}
\caption{The quantitative results on point-based generation tasks. }
\label{table:point_recon}

\end{table}

\noindent\textbf{Point Cloud Conditioned CAD Generation}.
The results are presented in \cref{table:point_recon}. Compared with the generation method DeepCAD~\cite{wu2021deepcad}, \methodname{} showcases significant improvements, \emph{i.e.}, \textbf{-4.47} Chamfer, \textbf{+15.51} F-score, and \textbf{+25.9} Normal C, demonstrating the better generation quality of \methodname{}. We contribute these improvements to two aspects: (1) autoregressive generation, which avoids the difficulty caused by joint modeling in DeepCAD; (2) learning on B-Rep, which supports freeform modeling, enhancing the generation performance on models with complex surfaces. Notably, our \methodname{} even outperforms the SOTA reconstruction method NVDNet~\cite{liu2024split} with \textbf{+0.90} F-score and \textbf{+1.12} Normal C, indicating the great potential of multimodal CAD generation methods.
In the qualitative comparison (Fig.~\ref{fig:pcd_cond_fig}), \methodname{} effectively generates the correct CAD model conditioned on the given point cloud, while DeepCAD~\cite{wu2021deepcad} fails on some instances, especially on instances with fine details. 

\noindent\textbf{Image Conditioned CAD Generation}.
To ensure a fair comparison, the input for \methodname{} and InstantMesh~\cite{xu2024instantmesh} is the same single image. The image is transparent to ensure methods can acquire information of multiple surfaces within one image. 
\cref{table:image_recon} shows the quantitative results. When compared with InstantMesh, our method shows clear advantages across all metrics, \emph{e.g.}, \textbf{+4.01} Chamfer, demonstrating that \methodname{} can generate CAD more accurately from the rendered CAD image. We suggest that, while InstantMesh is a general-purpose 3D mesh generator capable of producing high-quality 3D assets from a single conditional image, it lacks he essential topological priors inherent to CAD models. In contrast, \methodname{} includes such priors in the cascaded autoregressive generation network to generate edges and surfaces progressively, enabling a more precise capture of the complex and detailed geometries in CAD designs.
\cref{fig:img_cond_fig} provides the qualitatively evaluation. We observe that \methodname{} is able to reconstruct precise CAD instances while InstantMesh fails short of producing smooth surfaces.

\noindent\textbf{Text Conditioned CAD Generation}.
Since there are no established metrics for the text conditioned CAD generation task, we compare \methodname{} with the open-sourced Michelangelo~\cite{zhao2023michelangelo} using human evaluations and SOTA MLLMs, \emph{i.e.}, GPT-4o~\cite{hurst2024gpt} and Qwen-2.5-vl-72B~\cite{bai2025qwen2}. Specifically, we first invite 10 CAD designers to evaluate pairs of text and the generated models, and then select the one that is more consistent with textual description and of greater quality. For MLLMs, we input a prompt consisting of the conditional test description and multi-view images of the generated CAD models from both methods into the MLLMs, and then ask MLLMs to make a judgment as to which image better matches the given text description. After test on 20 models, the win-rate is presented in \cref{table:text_generate}. Our method achieves higher scores on both human and MLLMs, indicating that \methodname{} is more capable of generating models that match the text description.

\begin{table}
\centering
\setlength\tabcolsep{3.5 pt}
\small
\resizebox{0.95\linewidth}{!}{
\begin{tabular}{lccc}
\toprule
Method  & Chamfer $\downarrow$       & F-score $\uparrow$ & Normal C $\uparrow$ \\ \midrule
InstantMesh~\cite{xu2024instantmesh}  &  6.18  &  84.71  & 52.72    \\
CMT & \textbf{2.17} & \textbf{92.93} & \textbf{70.14} \\
\bottomrule
\end{tabular}
}
\caption{The quantitative results on image-based generation tasks.}
\label{table:image_recon}
\end{table}


\begin{table}[t]
  \centering
  \setlength\tabcolsep{7.5 pt}
\small
\resizebox{0.95\linewidth}{!}{
    \begin{tabular}{cccc}
    \toprule
    \multirow{2}[0]{*}{Method} & \multicolumn{3}{c}{Win rate \%} \\ 
          & GPT4o & Qwen  & Human \\  \midrule  
    Michelangelo~\cite{zhao2023michelangelo} & 25    & 35    & 20 \\
    \methodname{}   & \textbf{75}    & \textbf{55}    & \textbf{80} \\
    \bottomrule
    \end{tabular}
    }
    \vspace{-0.1em}
\caption{The results on text-conditioned generation tasks. We evaluate the generated models by multimodal large language models.}
\label{table:text_generate}
\vspace{-0.5em}
\end{table}



\begin{table}[t]
  \centering
  \setlength\tabcolsep{3.5 pt}
  \small
  \resizebox{0.99\linewidth}{!}{
    \begin{tabular}{cccccc}
    \toprule
    Cascade & Sampling steps & COV $\uparrow$  & MMD$\downarrow$   & JSD$\downarrow$  & Valid$\uparrow$ \\ \midrule  
    $\checkmark$ & 64/32 &   \textbf{75.71}  & \textbf{0.92}  &   \textbf{1.02} & \textbf{70.13} \\ 
          &  64+32    &   65.80  & 1.12 &  4.71  &  47.17 \\ \midrule
    $\checkmark$ & 64/32     &   \textbf{75.71}  & \textbf{0.92}  &   \textbf{1.02} & \textbf{70.13} \\ 
    $\checkmark$ & 32/16 &   74.97  & 1.08 &  1.08  &  67.93 \\
    $\checkmark$ & 16/8 &   70.26  & 1.16 &  2.10  &  47.80 \\
    $\checkmark$ & 8/4 &   N.A.  & N.A. &  N.A.  &  1.20 \\
    $\checkmark$ & 1/1 &   N.A.  & N.A. &  N.A.  &  0.10 \\
    \bottomrule
    \end{tabular}
    }
    \caption{Comparison of using cascade network and single network. X/Y: X sampling steps for edge generation and Y sampling steps for surface generation. X+Y: X+Y sampling steps for single network. N.A.: Unable to meet the number of models required for the Metrics because of low Valid ratio.}
  \label{tab:ablation}
  \vspace{-0.5em}
\end{table}

\subsection{Ablation Study}
\label{sec:ablation}

To investigate the effectiveness of key components of \methodname{}, we conduct ablation study on DeepCAD~\cite{wu2021deepcad}.

\noindent\textbf{Effectiveness of cascade network.}
To demonstrate the importance of using a cascade network instead of a single network for B-Rep generation, we ablate the cascade autoregressive generation network by comparing the single autoregressive generation network to generates edges and surfaces synchronously. The results are presented in \cref{tab:ablation}. The cascade network bring remarkable performance gains with \textbf{+9.91\%} \emph{Coverage}  and \textbf{+22.96\%} \emph{Valid}, which illustrates that the ``edges-then-surfaces" generation paradigm is critical for B-Rep generation. Applying cascade network, when generating surfaces, \methodname{} can leverage the information of edges, which reduces the difficulty of generation and thus enhances the performance.



\noindent\textbf{Effectiveness of autoregressive generation.}
To illustrate the importance of autoregressive generation, we ablate the sampling steps during generation. Specifically, we reduced the sampling steps of \methodname{} from sequence length 64/32 (\emph{i.e.}, one token at one step) to 32/16, 16/8, 8/4 or even 1/1 (\emph{i.e.}, generation all tokens at one step). As shown in \cref{tab:ablation}, the difficulty of modeling can be effectively reduced with the increase of sampling steps, thus improving the \emph{Valid} rate and diversity of generated models.

\section{Conclusion}
\label{sec:conclusion}

We introduce \methodname{}, the first multimodal CAD generation framework based on B-Rep, supporting both unconditional generation and conditonal generation tasks. The goal of \methodname{} is to achieve accurate CAD generation while conveniently processing users' multimodal instructions. To solve this challenging task, we propose to generate it through a cascade autoregressive generation network to enhance both accuracy and usability. Since there are no existing multimodal B-Rep datasets, to develop \methodname{}, we construct \datasetname{}, the largest CAD dataset with multimodal descriptions.
Extensive experiments including conditional and unconditional CAD generation demonstrate that our \methodname{} outperforms the existing methods with considerable improvement. Future work will extend our framework to more challenging tasks, such as multi-body generation, model completion, to further improve the applicability of our framework.

\section*{Acknowledgements}

\begingroup\raggedright This work was supported by the National Key R\&D Program of China(2022ZD0160201), Shanghai Artificial Intelligence Laboratory, the JC STEM Lab of AI for Science and Engineering, funded by The Hong Kong Jockey Club Charities Trust, the Research Grants Council of Hong Kong (Project No. CUHK14213224).  \par \endgroup
{
    \small
    \bibliographystyle{ieeenat_fullname}
    \bibliography{main}
}


\end{document}